\documentclass[10pt,twocolumn,letterpaper]{article}

\usepackage{fancyhdr}
\usepackage{cvpr}
\usepackage{times}
\usepackage{graphicx}
\usepackage{amsmath}
\usepackage{amssymb}
\usepackage{textcomp}
\usepackage{url}
\usepackage{booktabs}
\usepackage{footmisc}
\usepackage{subfig}
\usepackage{array,multirow}
\usepackage{rotating}
\usepackage[table]{xcolor}
\usepackage{xcoffins}
\usepackage{siunitx}
\usepackage[export]{adjustbox}
\NewCoffin\tablecoffin
\NewDocumentCommand\Vcentre{m}
{%
	\SetHorizontalCoffin\tablecoffin{#1}%
	\TypesetCoffin\tablecoffin[l,vc]%
}
\usepackage{tikz}
\usepackage{pgfplots}
\usepackage{pgfkeys}
\usetikzlibrary{pgfplots.groupplots}
\usetikzlibrary{external}

\newenvironment{customlegend}[1][]{%
	\begingroup
	\csname pgfplots@init@cleared@structures\endcsname
	\pgfplotsset{#1}%
}{%
\csname pgfplots@createlegend\endcsname
\endgroup
}%

\def\addlegendimage{\csname pgfplots@addlegendimage\endcsname}
\newcommand{\cmmnt}[1]{}
\definecolor{myteal}{RGB}{0,255,255}
\definecolor{mymag}{RGB}{255,0,255}
\definecolor{myblue}{RGB}{0,0,255}
\definecolor{myred}{RGB}{255,0,0}
\definecolor{mygreen}{RGB}{0,255,0}
\definecolor{myyellow}{RGB}{255,255,0}

 \cvprfinalcopy 

\ifcvprfinal\fi

\begin{document}
\pagestyle{fancy}
\title{Overhead Detection: Beyond 8-bits and RGB}

\author{Eliza Mace\textsuperscript{1}
	\and
	Keith Manville\textsuperscript{1}
	\and
	Monica Barbu-McInnis\textsuperscript{1}
	\and
	Michael Laielli\textsuperscript{2}
	\and
	Matthew Klaric\textsuperscript{2}\\
	{\tt\tiny\textsuperscript{1}MITRE, \{emace,kmanville,monicabm\}@mitre.org}
	\and
	Samuel Dooley\textsuperscript{2}\\
	{\tt\tiny\textsuperscript{2}NGA, \{michael.j.laielli,matthew.n.klaric,samuel.w.dooley\}@nga.mil}
}
\maketitle
\thispagestyle{fancy}
\begin{abstract}
This study uses the challenging and publicly available SpaceNet dataset to establish a performance baseline for a state-of-the-art object detector in satellite imagery. Specifically, we examine how various features of the data affect building detection accuracy with respect to the Intersection over Union metric.  We demonstrate that the performance of the R-FCN detection algorithm on imagery with a 1.5 meter ground sample distance and three spectral bands increases by over 32\% by using 13-bit data, as opposed to 8-bit data at the same spatial and spectral resolution. We also establish accuracy trends with respect to building size and scene density. Finally, we propose and evaluate multiple methods for integrating additional spectral information into off-the-shelf deep learning architectures. Interestingly, our methods are robust to the choice of spectral bands and we note no significant performance improvement when adding additional bands. 
\end{abstract}

\section{Introduction}
\begin{figure*}
	\begin{center}
		\includegraphics[width=0.8\linewidth]{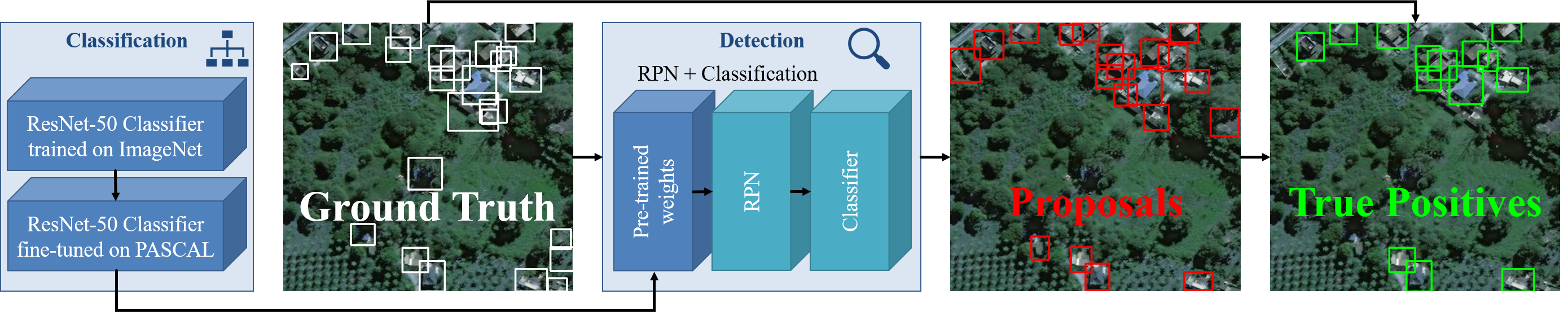}
	\end{center}
	\caption{The object detection architecture used in this study consists of two steps. First, weights pre-trained on a classification task are used to initialize a region proposal network that selects candidate regions from an image. Second, the proposed regions are classified during a separate classification step. All imagery in this figure is from DigitalGlobe.}
	\label{fig:pipeline}
\end{figure*}
As a result of recent investments by entities within the government and private sectors, an increased number of high resolution and multispectral satellites provide a massive amount of imagery for a wide range of applications, such as environmental monitoring, geographical hazard detection, land use and land cover mapping, urban planning, human rights related documentation, monitoring of conflict in inaccessible or prohibited areas and assessment of corporate growth. Identifying objects within these images is a natural way to quantitatively monitor events for each of these applications. With that said, object detection in satellite imagery presents challenges including large differences in visual appearances of objects caused by look-angle variation, few pixels on target, cluttered scenes, varying illumination and atmospheric conditions, and large object footprints (on the order of hundreds of square kilometers). In addition to exploring the impact of these challenges, our findings suggest that the unique features of overhead data, mainly the higher dynamic range, which can often be orders of magnitude higher than the range of typical consumer ground-level images, and additional spectral bands, eight bands versus the typical red, green, and blue channels, contribute to comparable object detection performance even when there is a three fold decrease in spatial resolution.

Inspired by the recent release of satellite image datasets like SpaceNet\footnote{https://aws.amazon.com/public-datasets/spacenet/}, we establish a baseline in terms of object size and scene density for state-of-the-art detectors when higher spectral resolution and higher bit-depth is available.  We devise an efficient way to augment the detection architecture to process multispectral data. Finally, we provide a trade space that describes detector performance related to spatial, spectral, and dynamic range resolution. 

This paper is organized as follows.  In Section 2, we discuss historical methods of automated information extraction from overhead images, as well as modern, deep learning-enabled object detection methods. We then compare and contrast these object detection algorithms in Section 3. Details of our baseline experiment, along with descriptions of our evaluation techniques and novel methods for expanding off-the-shelf architectures to accept data with additional spectral channels, are outlined in Section 4. Finally, we provide both a detailed results section, Section 5, and a brief conclusion, Section 6, that enumerate our findings and impacts on the larger research community.
\section{Related work}
\label{sec:related}
Efforts to obtain effective automated object detection algorithms on overhead images have a long history that extends at least several decades \cite{aviad1988road,quam1978road}. An early approach is to search for objects that best match a given template. For example, McKeown et. al \cite{mckeown1988cooperative} proposed a road tracking algorithm that measured how well a given object matches a target with respect to several parameters, such as width and contrast.  Another common early technique is Knowledge-Based Object Detection, which relies on hand-crafted rules for detection \cite{trinder1998knowledge}. One primary advantage of these models is that they require significantly less training data than modern deep learning methods. On the other hand, they tend to have relatively poor performance in terms of accuracy.  

As the spatial resolution of aerial imagery improved, neighborhoods of pixels could be analyzed for object detection. Object-Based Image Analysis does this, where the first step is to identify homogeneous groups of pixels from the image.  Then scene specific features (\eg texture, contextual semantic and geometric information) are extracted for classification by a support vector machine or some other classifier \cite{blaschke2010object,blaschke2014geographic}.  This is in line with the general approach of modern methods, which is to identify object proposals from the image and extract features for the proposals, and lastly apply a machine learning classifier on those features.  Modern machine learning techniques use convolutional neural networks (CNNs), where object proposals are learned from abstract features of the image that are also used for classification.  The models that use CNNs require significantly more training data than traditional feature extractors.  However, due to their impressive accuracy, and the recent availability of large-scale, annotated satellite imagery, we will focus on models with CNNs.  

Analyses of these models in terms of various parameters (\eg test-time speed, accuracy, and memory) on the ImageNet \cite{ImageNetpaper} and Microsoft Common Objects in Context (MS-COCO) \cite{COCO} datasets have been investigated previously \cite{canziani2016analysis,DBLP:journals/corr/HuangRSZKFFWSG016}. However, performance studies are lacking when the objects of interest are obscured, due to shadows, atmospheric conditions, or poor spatial resolution, as is often the case with satellite imagery. Multiple studies demonstrate the ability to improve the image quality of digital cell phone pictures taken in low-light conditions by using a higher dynamic range \cite{hasinoff2016burst,martinec2008noise,reinhard2010high}. Additionally, a multispectral approach to object detection was successfully used in \cite{mccool2016visual} and \cite{sa2016deepfruits} to detect fruits in an orchard, because a single sensor modality can rarely provide enough information to differentiate objects given varying illumination conditions, appearances and partial occlusions. Therefore, the natural question is whether a higher dynamic range, innately present in overhead sensors, and the additional spectral information will also yield better detection results for satellite imagery.    
\section{Detector architectures}
\label{sec:architectures}
Current state-of-the-art object detection systems fall into two categories: two-stage architectures that consist of region proposal generation followed by a separate, per-region classification (as done by the Faster Region-based CNN (Faster R-CNN) \cite{frcnn} and Region-based Fully Convolutional Network (R-FCN) \cite{rfcn} algorithms); and one-step systems that generate detections directly from image features (such as the Single Shot MultiBox Detector (SSD) \cite{DBLP:journals/corr/LiuAESR15} and the You Only Look Once (YOLO) approach \cite{YOLOpaper}). Both categories of detectors are built from a foundation of common network architectures such as (VGG \cite{vggpaper} or ResNet \cite{resnetpaper}) trained on ImageNet with detection mechanisms added to the network which produce bounding boxes and confidence scores.

A comprehensive comparison of Faster R-CNN, R-FCN, and SSD in terms of both speed and accuracy for the MS-COCO dataset was released in 2017 \cite{DBLP:journals/corr/HuangRSZKFFWSG016}. The study showed that the two-stage architectures, Faster R-CNN and R-FCN, provided higher accuracy, while SSD, being only one step, required less time to complete its detection tasks. Because our study aims to create an environment in which to compare accuracies across detection tasks on varying bit-depths, and spatial and spectral resolutions, we elected to use a two-stage object detection scheme for its higher accuracy results. 

To select between the Faster R-CNN algorithm and R-FCN algorithms, we compared their architectures and recent performance on non-overhead tasks. Both algorithms eliminate the need for external object proposals by introducing a Region Proposal Network (RPN) that learns anticipated regions, known as regions of interest (ROIs), from CNN features, as shown in Figure \ref{fig:pipeline}. In the Faster R-CNN framework, these features are pushed upstream to an object detection network \cite{Yangpaper}. In contrast, R-FCN classifies the ROIs generated by its RPN as either desired objects or background by using position sensitive score maps generated by fully convolutional networks. R-FCN addresses location variance by proposing different object regions and location invariances by having each region proposal refer back to the same score map. Being fully convolutional makes R-FCN much faster than Faster R-CNN with comparable performance to the leading detector architectures \cite{rfcn,DSSDpaper}. Additionally, Dai \etal \cite{rfcn} observed a boost in R-FCN performance by incorporating online hard example mining (OHEM) during the forward pass at training by evaluating the loss of all proposed ROIs, selecting the ones with the highest loss, and performing backpropagation on the selected set. Given the computational efficiency and performance edge, we use R-FCN with OHEM throughout our baseline analysis of overhead data.  
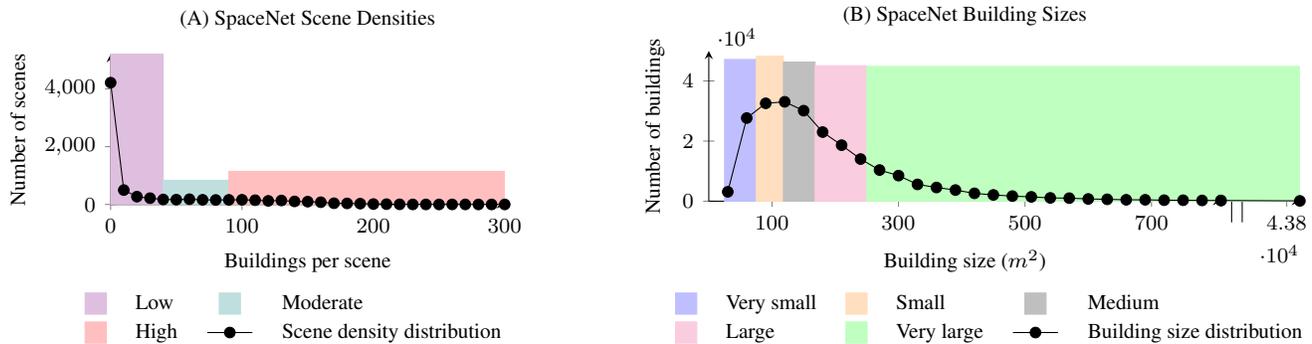
\begin{figure*}
	\centering
		\begin{tikzpicture}[font=\footnotesize]
		\begin{axis}[
		title=(A) SpaceNet Scene Densities,
		xlabel={Buildings per scene},
		ylabel={Number of scenes},
		axis x line=bottom, 
		axis y line=left,
		ymin=0,
		xmin=0,
		xmax=300,			
		scale only axis,
		width=0.3\linewidth,
		height=2cm,
		clip mode=individual
		]
		\addplot+ [ybar interval,color=violet!25,mark=none,fill=violet!25] coordinates{
			(0,5205) (40,0) };
		
		\addplot+ [ybar interval,color=teal!25,mark=none,fill=teal!25] coordinates{
			(40,835) (90,0)};
		
		\addplot+ [ybar interval,color=red!25,mark=none,fill=red!25] coordinates{
			(90,1146) (300,1146)};
		
		\addplot[color=black,mark=*,mark options={fill=black}]coordinates{
			(0,4217) (10,498) (20,270) (30,218) (40,172) (50,166) (60,181) (70,161) (80,155) (90,159) (100,162) (110,147) (120,124) (130,137) (140,105) (150,97) (160,77) (170,41) (180,36) (190,28) (200,14) (210,10) (220,4) (230,1) (240,1) (250,0) (260,1) (270,1) (280,0) (290,0) (300,1)
		};
		
		\end{axis}
		\end{tikzpicture}
		\hfill
	\begin{tikzpicture}[font=\footnotesize]
	\begin{groupplot}[
	group style={
		group name=axisbreak, group size=2 by 1, xticklabels at=edge bottom,horizontal sep=0pt},
	height=2cm,xmin=0,xmax=43800]
	
	\nextgroupplot[
	title=(B) SpaceNet Building Sizes,
	xlabel={Building size ($m^2$)},
	ylabel={Number of buildings},
	axis x line=bottom, 
	axis y line=left,
	xmin=0,
	xmax=810,
	ymin=0,
	ymax=50000,	
	xtick={100,300,500,700},		
	scale only axis,
	width=0.39\linewidth,
	height=2cm,
	clip mode=individual,
	ylabel near ticks
	]
	\addplot+ [ybar interval,color=blue!25,mark=none,fill=blue!25] coordinates{
		(25,47136) (75,0) };
	
	\addplot+ [ybar interval,color=orange!25,mark=none,fill=orange!25] coordinates{
		(75,48212) (118,0)};
	
	\addplot+ [ybar interval,color=black!25,mark=none,fill=black!25] coordinates{
		(118,46294) (168,0)};
	
	\addplot+ [ybar interval,color=magenta!25,mark=none,fill=magenta!25] coordinates{
		(168,45052) (250,0)};
	
	\addplot+ [ybar interval,color=green!25,mark=none,fill=green!25] coordinates{
		(250,44836) (810,0)};
	
	\addplot[color=black,mark=*,mark options={fill=black}]coordinates{
		(30,3117) (60,27651) (90,32553) (120,33058) (150,30110) (180,22989) (210,18633) (240,14015) (270,10359) (300,8506) (330,5583) (360,4555) (390,3704) (420,2609) (450,2086) (480,1693) (510,1434) (540,1089) (570,1009) (600,746) (630,621) (660,464) (690,461) (720,327) (750,314) (780,250) (810,238)};
	
	\nextgroupplot[
	xmin=43740,
	xmax=43800,
	ymin=0,
	ymax=50000,
	xtick={43790},
	axis x line=bottom,
	y axis line style={draw opacity=0},
	hide y axis,
	axis x discontinuity=parallel,
	yticklabels={},
	scale only axis,
	width=0.06\linewidth,
	height=2cm,
	clip mode=individual
	]
	
	\addplot [ybar interval,color=green!25,mark=none,fill=green!25] coordinates{
		(43740,44836) (43800,0)};
	\addplot [color=black,mark=*,mark options={fill=black}]coordinates{
		(43740,238) (43800,100)};
		
	\end{groupplot}
	\end{tikzpicture}
	\begin{tikzpicture}[font=\footnotesize]
		\hspace{1cm}
	\begin{customlegend}[legend columns=2,legend style={align=left,draw=none,column sep=2ex},legend cell align=left,legend entries={Low, Moderate, High, Scene density distribution}]
	\addlegendimage{color=violet!25,only marks, mark=square*,mark options={scale=2,fill=violet!25}}
	\addlegendimage{color=teal!25,only marks, mark=square*,mark options={scale=2,fill=teal!25}}
	\addlegendimage{color=red!25,only marks, mark=square*,mark options={scale=2,fill=red!25}}
	\addlegendimage{color=black, mark=*,mark options={fill=black}}
	\end{customlegend}
	\end{tikzpicture}
	\hfill
	\begin{tikzpicture}[font=\footnotesize]
	\begin{customlegend}[legend columns=3,legend style={align=left,draw=none,column sep=2ex},legend cell align=left,legend entries={Very small, Small, Medium, Large, Very large, Building size distribution}]
	\addlegendimage{color=blue!25,only marks, mark=square*,mark options={scale=2,fill=blue!25}}
	\addlegendimage{color=orange!25,only marks, mark=square*,mark options={scale=2,fill=orange!25}}
	\addlegendimage{color=black!25,only marks, mark=square*,mark options={scale=2,fill=black!25}}
	\addlegendimage{color=magenta!25,only marks, mark=square*,mark options={scale=2,fill=magenta!25}}
	\addlegendimage{color=green!25,only marks, mark=square*,mark options={scale=2,fill=green!25}}
	\addlegendimage{color=black, mark=*,mark options={fill=black}}
	\end{customlegend}
	\end{tikzpicture}
	\caption{(A) Distribution of building sizes in SpaceNet and corresponding subset bins and (B) distribution of per-scene building density subset bins used in this study.}
	\label{fig:densitydist}
	\label{fig:sizedist}
\end{figure*}
\section{Experimental setup}
\label{sec:experiment}
\subsection{Data}
\label{sec:data}
SpaceNet includes WorldView-2 imagery data of two types: 3-band images, at approximately 0.5 meter ground sample distance (GSD) at 8-bits; and 8-band multispectral images, at approximately 1.5 meter GSD at 13-bits, as well as corresponding building footprint labels. The WorldView-2 sensor collects 11-bit imagery, but due to post-processing, the publicly available SpaceNet imagery has pixels as deep as 13 bits. From here on, we refer to the 0.5 m GSD images as ``high resolution'' (HR) and the 1.5 m GSD images as ``low resolution'' (LR). The 3-band images are standard color images where the three channels reflect light around the 659, 546, and 478 nanometer (nm) wavelengths, corresponding to red, green, and blue channels.  The 8-band imagery includes additional spectral bands for the coastal blue, yellow, red edge, near infrared 1 (NIR1) and near infrared 2 (NIR2) channels, which correspond to the center wavelengths of 427, 608, 724, 883, and 949 nm, respectively. Our results are based on Area of Interest 1, covering scenes with buildings across Rio de Janeiro, Brazil. The complete scenes of each data type are broken down into over 7,000 individual scenes, each approximately $45,000m^2$. Figures~\ref{fig:sizedist} (A) and (B) indicate the distribution of buildings per scene (scene density) and building sizes across the dataset, respectively, along with categorical data bins. These categories were established for analysis as well as to put into perspective the wide variation across the scenes and within the building object class itself. We randomly assigned 80\% of the scenes to a training set and reserved the remaining 20\% of the images for testing. 

\begin{figure}
	\begin{center}		
		\includegraphics[width=0.6\linewidth]{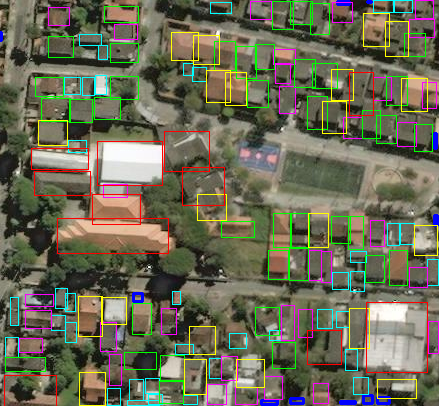}
		\begin{tikzpicture}[font=\footnotesize]
		\begin{customlegend}[legend columns=3,legend style={align=left,draw=none,column sep=2ex},legend cell align=left,legend entries={Discard, Very small, Small, Medium, Large, Very large}]
		\addlegendimage{color=myblue, only marks,thick,mark=square,mark options={scale=2}}
		\addlegendimage{color=myteal, only marks,mark=square,mark options={scale=2}}
		\addlegendimage{color=mymag,only marks,mark=square,mark options={scale=2}}
		\addlegendimage{color=mygreen, only marks,mark=square,mark options={scale=2}}
		\addlegendimage{color=myyellow, only marks,mark=square,mark options={scale=2}}
		\addlegendimage{color=myred, only marks,mark=square,mark options={scale=2}}
		\end{customlegend}
		\end{tikzpicture}
	\end{center}
	\caption[Example of SpaceNet building size]{Example of SpaceNet building sizes corresponding to our subset bins. All imagery in this figure is from DigitalGlobe.}
	\label{fig:boundingboxes}	
\end{figure}

The building footprint labels amount to about 300,000 bounding boxes. Of these boxes, those with an area less than $25m^2$ were discarded, which amounted to approximately 5\% of the total annotations. Qualitative inspection revealed that many of these eliminated annotations were created when buildings were split across scene divides. Some discarded examples are indicated in Figure \ref{fig:boundingboxes} by bold, dark blue boxes.

As mentioned above, the SpaceNet dataset's annotations were originally in the form of per-pixel building footprints, as opposed to bounding boxes. Therefore, the bounding boxes generated from the footprint annotations were precisely fit around the top-, bottom-, left-, and right-most pixels of the buildings themselves. We noticed that this created an inherent difference in the content captured by bounding boxes depending upon the orientation of the spanned building itself. Specifically, if a building was oriented such that its edges were parallel to the edges of the scene on which it appeared, it contained virtually no background content, as opposed to buildings at an angle within their scene whose boxes contained increasing amounts of background as their alignment deviated from the edges of the scene. In order to alleviate this discrepancy, and inspired by \cite{Yangpaper} and \cite{COWCpaper}, who each noted an accuracy improvement when contextual background was increased, we included $6m$ padding around all bounding boxes. The amount of padding was chosen empirically not only to ensure that the aforementioned motivations where addressed, but also so that the increased bounding box area would not artificially inflate the Intersection over Union (IoU) score during analysis (especially for small bounding boxes whose overall area is increased by a higher percentage when padding is added).
\subsection{Evaluation metrics}
We use an $F_1$ Score, the harmonic mean of precision and recall, to evaluate and compare detector performance. As the $F_1$ Score requires a count of true detections, each proposed bounding box is evaluated to determine whether it could be counted as such using an algorithm detailed by Russakovsky \etal \cite{ImageNetpaper}. For each candidate box that was assigned a confidence score over a predetermined threshold, if that candidate is the closest of any of the proposed bounding boxes to a ground truth box as measured by an IoU score, and that IoU score is greater than a specified threshold, then the candidate is counted as a true positive, and the corresponding ground truth box is no longer considered as a possible match for any other candidate. Proposed boxes that do not meet the aforementioned criteria were considered to be false positive detections. For the purposes of counting false positives when analyzing the effect of object size on performance, we computed the area of each errant detection box and assigned it a size category based on the original annotation size categories.
\subsection{Approach}
Although ResNet-101 is the top performer described in \cite{resnetpaper} and \cite{DSSDpaper} on PASCAL VOC \cite{pascal} and MS-COCO datasets, an initial investigation with overhead data showed that ResNet-50 provides a 15\% increase in recall over its 101-layer counterpart. As a result, we utilize the ResNet-50 model as a feature extractor, pre-trained on the PASCAL VOC 2012 dataset in our study. The detector uses $\textit{Argmax}$ matching between anchors and ground truth locations, typical box encoding and a smooth $L_1$ loss function. The detector is trained with stochastic gradient descent with momentum of 0.9. The R-FCN minibatch size for RPN training is set to 256, while the minibatch size for the box classifier training is 64. The learning rate is set to 0.001 and we train end-to-end on the SpaceNet data for up to 60,000 iterations. 

All experiments were performed in Caffe \cite{jia2014caffe} on K80 GPUs using the open source implementations of R-FCN\footnote{https://github.com/daijifeng001/R-FCN}. It is important to note that we changed Caffe methods that performed an ``image read'' to call libtiff\footnote{http://www.libtiff.org/}, a library equipped to handle images with higher bit-depths.
\subsection{Network expansion for multispectral data}
\begin{figure*}
	\begin{center}
		\includegraphics[width=0.8\linewidth]{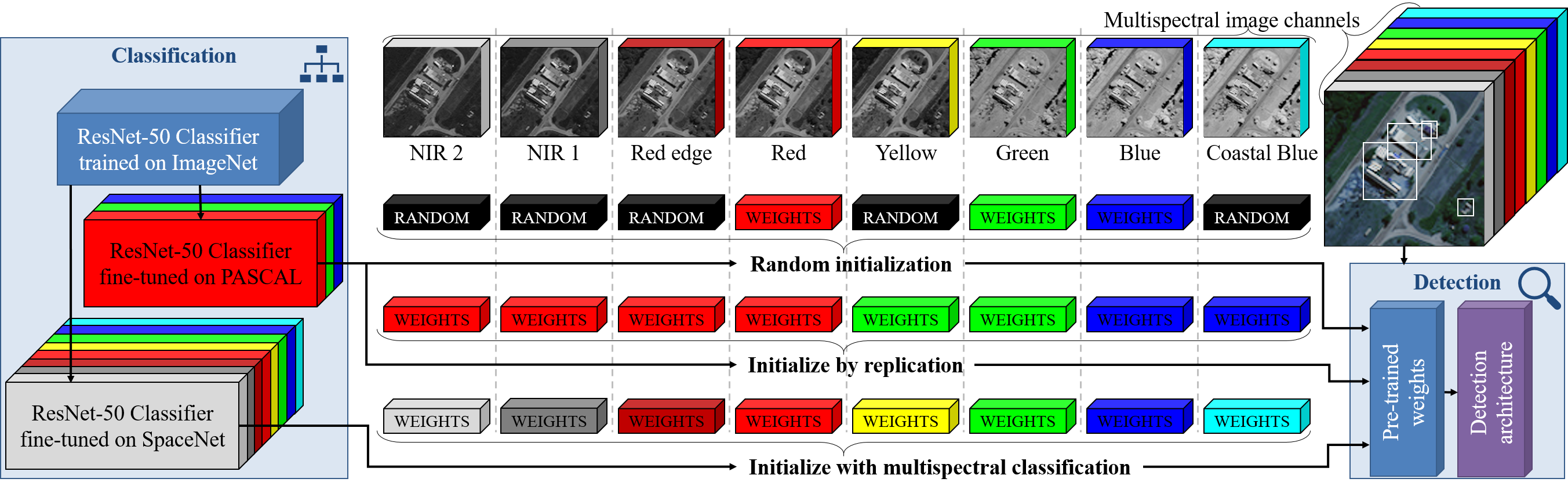}
	\end{center}
	\caption{Allocation of learned 3-band weights with random, replicated, or multispectral classification weights to initialize 8-band multispectral data. All imagery in this figure is from DigitalGlobe.}
	\label{fig:weightallocation}
\end{figure*}
The architecture describing most, if not all, state-of-the-art detectors supports mainly RGB datasets; to detect objects in multispectral images, we have to modify said architecture. A careful look at the network architecture shows that we only have to change the parameters in the first convolutional layer to accept image data with more channels. For example, for the ResNet-50 architecture, the shape of the data, $1\times3\times224\times224$, becomes $1\times8\times224\times224$ and the weight parameters in ``conv1'' become $64\times8\times7\times7$ in the 8-band instance as opposed to $64\times3\times7\times7$ in the 3-band, RGB, case.

Once the network architecture can accommodate multispectral datasets, the question of how best to initialize the weights becomes apparent.  Mainly, we aim to initialize the weights for the additional five bands while keeping the three RGB weights intact. We consider three ways to initialize the network: random initialization, initialization by replication, and initialization with multispectral classification.

We consider random weight initialization to be the simplest option (see Figure \ref{fig:weightallocation}).  However, knowing that performance improves when a network is fine-tuned on a larger and similar dataset \cite{finetunepaper,dataDepInit}, we speculated that a superior option would be to replicate the learned weights from the data specific RGB network to the remaining bands in the multispectral dataset. Considering that the multispectral bands are somewhat correlated in visual structure, allocating the 3-band weights to the remaining five bands, as shown in Figure \ref{fig:weightallocation}, should, intuitively, improve performance. To test this hypothesis and select which of these two initialization methods should be used for further experiments, we performed a preliminary experiment to compare their respective training progress, the results of which are shown in Figure \ref{fig:initializationcomparison}. Based on this result, we proceed with the method of replicating the 3-band weights across the uninitialized bands for training, as this method outperformed random initialization, per our expectation.
\begin{figure}[h]
	\centering
	
	\begin{tikzpicture}[font=\footnotesize]
	\begin{axis}[
	title=8-band Network Initialization,
	xlabel={Training iterations (in thousands)},
	ylabel={$F_1$ Score},
	axis x line=bottom, 
	axis y line=left,
	legend cell align=left,
	ymin=0,
	ymax=0.2,
	xmin=0,
	xmax=70,
	yticklabel style={
		/pgf/number format/fixed,
		/pgf/number format/precision=2,
		/pgf/number format/fixed zerofill
	},
	scaled y ticks=false,
	legend pos=south east,
	legend style={draw=none},
	scale only axis,
	width=0.8\linewidth,
	height=2.25cm
	]
	\addplot+[color=blue,mark=*,mark options={fill=blue}]
	coordinates{
		(10,0.110)
		(20,0.122)
		(30,0.141)
		(40,0.152)
		(50,0.149)
		(60,0.153)
	};
	\addplot+[color=red,mark=*,mark options={fill=red}]
	coordinates{
		(10,0.161)
		(20,0.176)
		(30,0.183)
		(40,0.185)
		(50,0.185)
		(60,0.190)
	};
	\addlegendentry{Random initialization}
	\addlegendentry{Initialize by replication}
	\end{axis}
	\end{tikzpicture}
	\caption[8-band network initialization comparison]{Preliminary comparison of 8-band initialization methods which demonstrates that using copies of the 3-band weights as a starting point for training outperforms randomly initializing the remaining bands' weights.\\ \tiny   }
	\label{fig:initializationcomparison}
	\vfill
	\vfill
	\begin{tikzpicture}[font=\footnotesize]
	\begin{axis}[
	title=Overall Detector Performance,
	xlabel={Metric},
	ylabel={Score},
	axis y discontinuity=parallel,
	xmin=0.7,
	xmax=3.5,
	ymin=0.15,
	ymax=0.47,
	axis x line=bottom, 
	axis y line=left,
	xtick={0.7,1.7,2.7},
	xticklabels={,,,},
	xmajorgrids,
	legend style={at={(0.5,-0.15)},anchor=north,draw=none,column sep=2ex},
	legend columns=2,
	legend cell align=left,
	scale only axis,
	width=0.8\linewidth,
	height=4.25cm,
	xlabel near ticks,
	xlabel shift = -1.5 ex
	]
	\addplot+[color=red, only marks, mark=diamond*,mark options={scale=1.5,fill=red},error bars/.cd, y dir=both, y explicit]
	coordinates{
		(1.00,0.345) +- (0.042,0.042)
		(2.00,0.421) +- (0.029,0.029)
		(3.00,0.308) +- (0.049,0.049)
	};
	\addlegendentry{8-bit HR 3-band}
	\node[] at (axis cs: 1.15,0.17) {$F_1$};
	\addplot+[color=blue, only marks, mark=*,mark options={scale=1.5,fill=blue},error bars/.cd, y dir=both, y explicit]
	coordinates{
		(1.15,0.270) +- (0.020,0.020)
		(2.15,0.330) +- (0.040,0.040)
		(3.15,0.249) +- (0.035,0.035)
	};
	\addlegendentry{8-bit LR 3-band}
	\node[] at (axis cs: 2.15,0.17) {Precision};
	\addplot+[color=green, only marks, mark=triangle*,mark options={scale=1.5,fill=green},error bars/.cd, y dir=both, y explicit]
	coordinates{
		(1.30,0.362) +- (0.018,0.018)
		(2.30,0.386) +- (0.037,0.037)
		(3.30,0.361) +- (0.032,0.032)
	};
	\addlegendentry{13-bit LR 3-band}
	\node[] at (axis cs: 3.15,0.17) {Recall};
	\addplot+[color=orange, only marks, mark=square*,mark options={scale=1.5,fill=orange},error bars/.cd, y dir=both, y explicit]
	coordinates{
		(1.45,0.350) +- (0.019,0.019)
		(2.45,0.392) +- (0.025,0.025)
		(3.45,0.332) +- (0.021,0.021)
	};
	\addlegendentry{13-bit LR 8-band}
	\end{axis}
	\end{tikzpicture}
	\caption{5-fold cross-validation results for R-FCN with OHEM shows statistically significant differences between low resolution, 8-bit data and all other data types. Error bars represent two standard deviations from the mean of the scores from the five trials.}
	\label{fig:cross}
\end{figure}

The third option for weight initialization explores the same concept of allocating existing weights to initialize all spectral bands to the pre-training phase of the detection pipeline. In contrast to the aforementioned method of initalization by replication, we were interested in obtaining unique weights for each of the eight bands by training the ResNet-50 classifier with 8-band imagery as a precursor to training the R-FCN end-to-end detection process with the 8-band data, hypothesizing that this additional training step would help capture visual features of the five non-RGB bands. Ideally, the classification task would be completely trained on 8-band data, but since such a volume of annotated multispectral data was not readily available, we relied on fine-tuning. To do this, we expanded the ResNet-50 weights trained on ILSVRC \cite{ImageNetpaper} by following the initialization by replication procedure shown in Figure \ref{fig:weightallocation}. Then, we trained this classifier using small patches extracted from our SpaceNet training scenes. The patches contained single buildings, as defined by the provided ground-truth bounding boxes with padding, as described in section \ref{sec:data}. We also created an equal number of patches that did not contain any part of a building to create a negative class, and fine-tuned the aforementioned ResNet architecture to do binary classification (i.e. ``building'' or ``not building''). Lastly, we trained R-FCN end-to-end on top of the weights from the fine-tuned ResNet-50 classifier to detect buildings from the 8-band dataset. During the training process, we ensured that all of the patches used to train the classifier were from the same scenes that were in the R-FCN training set, so that none of the objects that were used for evaluation in the testing process had been seen before by the classifier or detector.
\section{Results}
\label{sec:results}
\subsection{Overall detector performance}
\label{sec:overall_results}
The precision, recall, and $F_1$ Scores for data with different combinations of bit-depth, and spatial and spectral resolutions are enumerated in Table \ref{tab:apples-to-apples}.  As expected, performance drops when spatial resolution decreases. Motivated by improvements seen in \cite{DBLP:journals/corr/LiuAESR15} and \cite{smallobjs} we also note that a simple spatial rescale of the low resolution imagery provides an order of magnitude increase in performance (see rows 2 and 3 of Table \ref{tab:apples-to-apples}). Furthermore, taking advantage of the entire bit-depth level available in the native LR imagery can improve performance to levels comparable to that of the HR, 8-bit data. This increase in performance for the rescaled, LR, 13-bit data, seems fairly robust to the choice of spectral bands. This observation was upheld by an additional experiment in which we tested a different combination of bands to create a false-color RGB image, namely the 427, 608, and 724 nm bands (see Table \ref{tab:apples-to-apples}, row 5). Furthermore, the additional spectral bands do not seem to contribute much to the overall performance. This was true irrespective of the initialization choice, initialization by replication or initialization with multispectral classification, where the difference between the scores was minimal. Based on this observation, we can hypothesize that the spatial features are much more dominant than the spectral features. 
\begin{table*}
	\centering
	\small
	\begin{tabular}{c c c l c c c}
		\toprule
		 \textbf{Spatial resolution} & \textbf{Image size (pixels)} & \textbf{Dynamic range (bits)} & \textbf{Spectral bands} & \textbf{$F_1$ Score} & \textbf{Precision} & \textbf{Recall} \\ \midrule
			HR  & $439\times406$ & 8 & 3 (RGB) & 0.37 & 0.44 & 0.34 \\
			LR  & $110\times102$ & 8 & 3 (RGB) & 0.03  & 0.11 & 0.02\\
			LR  & $550\times510$ & 8 & 3 (RGB) & 0.28 & 0.35 & 0.25 \\
			LR  & $550\times510$ & 13 & 3 (RGB) & 0.37 & 0.41 & 0.35 \\
			LR & $550\times510$ & 13 & 3 (False color) & 0.37 & 0.42 & 0.34 \\
			LR  & $550\times510$ & 13 & 8 -- Replication& 0.36 & 0.39 & 0.34  \\
			LR & $550\times510$ & 13 & 8 -- Multispectral class. & 0.34 & 0.39 & 0.32 \\
		\hline
	\end{tabular}
    \caption[Detector performance comparison]{Bit depth and spectral variation performance at IoU 0.5 and detection confidence 0.5.}
    \label{tab:apples-to-apples}
\end{table*}

To explore whether some of the observed differences in performance were statistically significant when the input data includes the higher dynamic range as well as additional spectral bands, we performed a five-fold cross-validation experiment. Five unique sets of training and testing images, each with an 80/20 ratio between the two sets, were evaluated under the exact same conditions using R-FCN with OHEM. Figure \ref{fig:cross} describes the cross-validation findings, where, as already seen, the 13-bit resized LR images exhibit the same performance or better than their HR counterparts. We quantified this observation using an unpaired t-test for a two-tailed hypothesis, resulting in a \emph{p}-value between the two datasets of 0.155, which indicates the difference is not statistically significant. In contrast, comparing the $F_1$ Scores of both the HR 8-bit and LR 13-bit to the LR 8-bit data yield values of \emph{p}-value $<0.00001$ and \emph{p}-value $=0.000096$, respectively, both of which indicate statistically significant differences.  
\begin{figure*}
	\begin{center}
		\begin{tikzpicture}	[font=\footnotesize]	
			\begin{axis}[
				scale only axis,
				width=0.25\textwidth,
				height=3.7cm,
				title=(A) Detector Performance by Scene Density,
				xlabel={Scene Density},
				ylabel={$F_1$ Score},
				xmin=0.7,
				xmax=3.5,
				ymin=0,
				ymax=0.75,
				axis x line=bottom, 
				axis y line=left,
				xtick={0.7,1.7,2.7},
				xticklabels={,,,},
				xmajorgrids,
				xlabel near ticks,
				xlabel shift = -1.5 ex
				]
			\addplot+[color=red, only marks, mark=diamond,error bars/.cd, y dir=both, y explicit]
				coordinates{
					(1.00,0.271) +- (0.042,0.042)
					(2.00,0.577) +- (0.049,0.049)
					(3.00,0.536) +- (0.048,0.048)
				};

			\node[] at (axis cs: 1.15,0.05) {Low};
			\addplot+[color=blue, only marks, mark=o,error bars/.cd, y dir=both, y explicit]
				coordinates{
					(1.15,0.197) +- (0.012,0.012)
					(2.15,0.481) +- (0.051,0.051)
					(3.15,0.469) +- (0.035,0.035)
				};

			\node[] at (axis cs: 2.2,0.05) {Moderate};
			\addplot+[color=green, only marks, mark=triangle,error bars/.cd, y dir=both, y explicit]
				coordinates{
					(1.30,0.286) +- (0.024,0.024)
					(2.30,0.592) +- (0.037,0.037)
					(3.30,0.561) +- (0.009,0.009)
				};
		
			\addplot+[color=orange, only marks, mark=square,error bars/.cd, y dir=both, y explicit]
				coordinates{
					(1.45,0.273) +- (0.022,0.022)
					(2.45,0.586) +- (0.021,0.021)
					(3.45,0.554) +- (0.017,0.017)
				};
			\node[] at (axis cs: 3.15,0.05) {High};				
			\end{axis}
		\end{tikzpicture}
		\begin{tikzpicture}[font=\footnotesize]			
			\begin{axis}[
				scale only axis,
				width=0.45\textwidth,
				height=3.7cm,
				title=(B) Detector Performance by Building Size,
				xlabel={Building Size},
				ylabel={$F_1$ Score},
				xmin=0.7,
				xmax=5.5,
				ymin=0,
				ymax=0.75,
				axis x line=bottom, 
				axis y line=left,
				xtick={0.7,1.7,2.7,3.7,4.7},
				xticklabels={,,,},
				xmajorgrids,
				xlabel near ticks,
				xlabel shift = -1.5 ex
				]
			\addplot+[color=red, only marks, mark=diamond,error bars/.cd, y dir=both, y explicit]
				coordinates{
					(1.00,0.244) +- (0.061,0.061)
					(2.00,0.513) +- (0.038,0.038)
					(3.00,0.597) +- (0.044,0.044)
					(4.00,0.656) +- (0.034,0.034)
					(5.00,0.642) +- (0.077,0.077)
				};
			\node[align=center] at (axis cs: 1.15,0.07) {Very\\Small};
			\addplot+[color=blue, only marks, mark=o,error bars/.cd, y dir=both, y explicit]
				coordinates{
					(1.15,0.200) +- (0.039,0.039)
					(2.15,0.437) +- (0.026,0.026)
					(3.15,0.507) +- (0.049,0.049)
					(4.15,0.560) +- (0.034,0.034)
					(5.15,0.555) +- (0.065,0.065)
				};
			\node[] at (axis cs: 2.15,0.05) {Small};
			\addplot+[color=green, only marks, mark=triangle,error bars/.cd, y dir=both, y explicit]
				coordinates{
					(1.30,0.316) +- (0.049,0.049)
					(2.30,0.527) +- (0.034,0.034)
					(3.30,0.582) +- (0.042,0.042)
					(4.30,0.655) +- (0.041,0.041)
					(5.30,0.709) +- (0.033,0.033)
				};
			\node[] at (axis cs: 3.2,0.05) {Medium};
			\addplot+[color=orange, only marks, mark=square,error bars/.cd, y dir=both, y explicit]
				coordinates{
					(1.45,0.303) +- (0.042,0.042)
					(2.45,0.517) +- (0.021,0.021)
					(3.45,0.581) +- (0.017,0.017)
					(4.45,0.654) +- (0.034,0.034)
					(5.45,0.699) +- (0.019,0.019)
				};
			\node[] at (axis cs: 4.15,0.05) {Large};
			\node[align=center] at (axis cs: 5.15,0.07) {Very\\Large};
			\end{axis}
		\end{tikzpicture}
		\begin{tikzpicture}[font=\footnotesize]
		\begin{customlegend}[legend columns=4,legend style={align=left,draw=none,column sep=2ex},legend cell align=left,legend entries={8-bit HR 3-band,8-bit LR 3-band,13-bit LR 3-band,13-bit LR 8-band}]
			\addlegendimage{color=red, only marks, mark=diamond}
			\addlegendimage{color=blue, only marks, mark=o}   
			\addlegendimage{color=green, only marks, mark=triangle}
			\addlegendimage{color=orange, only marks, mark=square}
		\end{customlegend}
	\end{tikzpicture}
	\end{center}
	\caption{5-fold cross-validation results for R-FCN with OHEM show a disparity between 8-bit, low resolution data and other data types exists across various image densities and building sizes. In both figures, the 8-band classifier results were derived from the detector with weights initialized by replication. All error bars show two standard deviations from the mean computed from our five-fold experiments.}
	\label{fig:density}
	\label{fig:size}
\end{figure*}
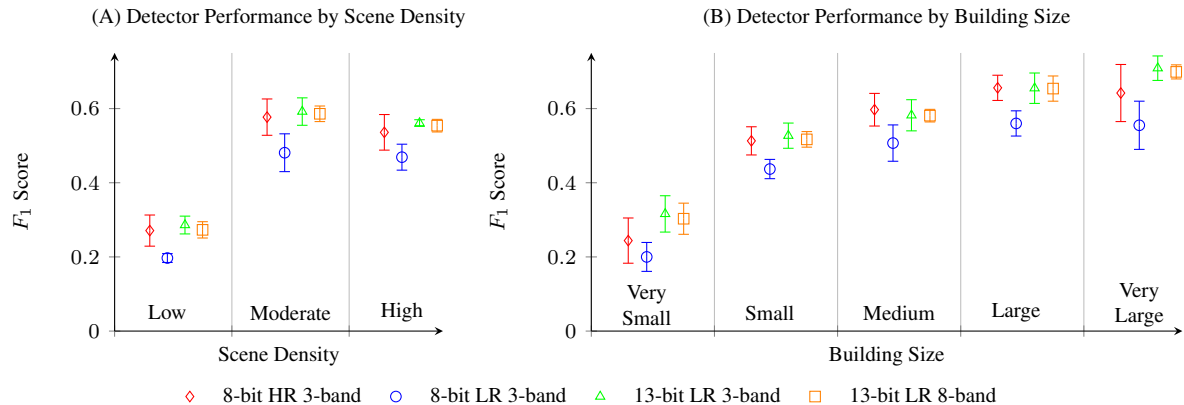
\subsection{Performance by scene density and object size}
\begin{figure*}[htbp]
	\centering
	\begin{tabular}{p{2.5cm} | c c c}
		\textbf{Data} & Low Density Scene & Moderate Density Scene & High Density Scene \\ \hline
		 8-bit HR 3-band & \Vcentre{\subfloat{\includegraphics[width = 1.6in]{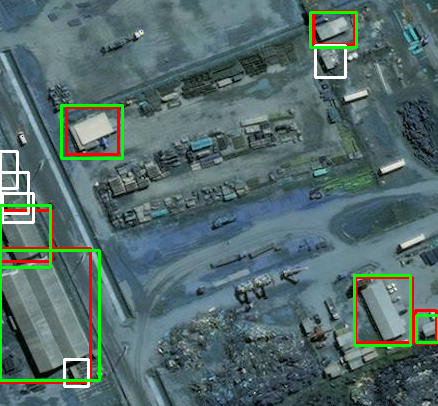}}} &
		\Vcentre{\subfloat{\includegraphics[width = 1.6in]{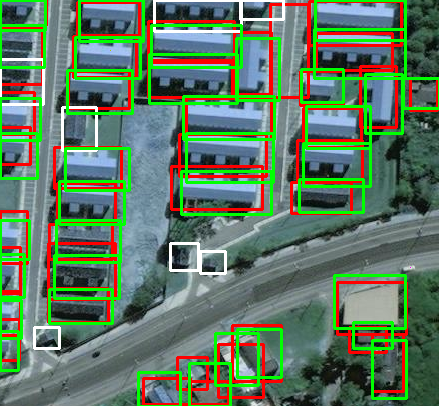}}} &
		\Vcentre{\subfloat{\includegraphics[width = 1.6in]{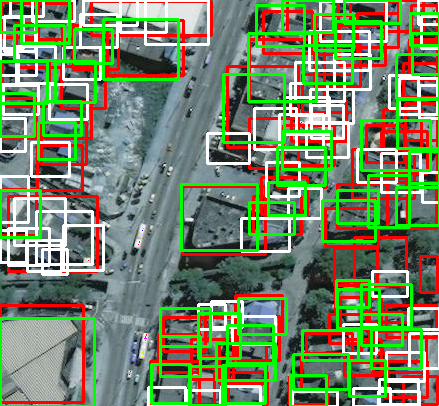}}}\\
		8-bit LR 3-band & \Vcentre{\subfloat{\includegraphics[width = 1.6in]{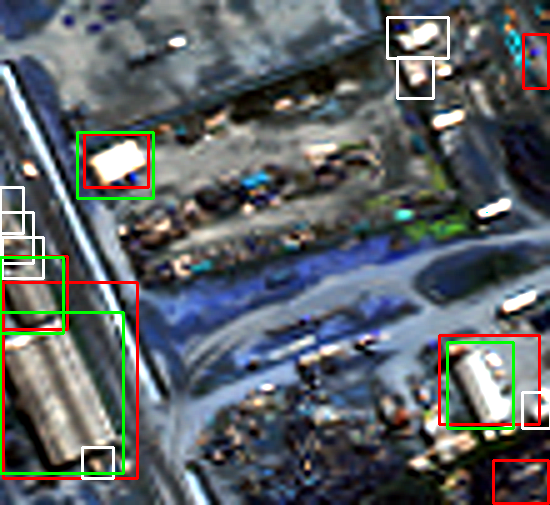}}} &
		\Vcentre{\subfloat{\includegraphics[width = 1.6in]{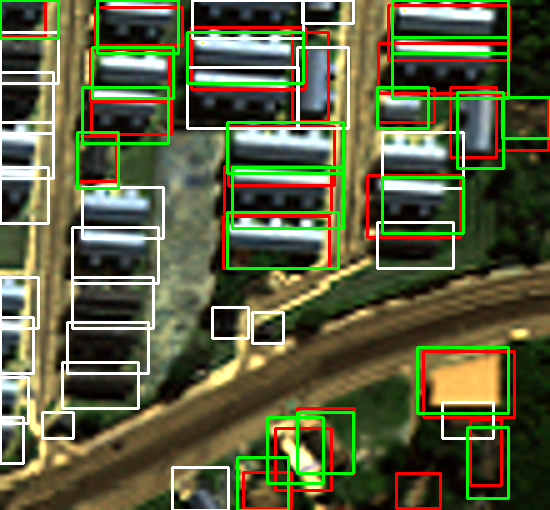}}} &
		\Vcentre{\subfloat{\includegraphics[width = 1.6in]{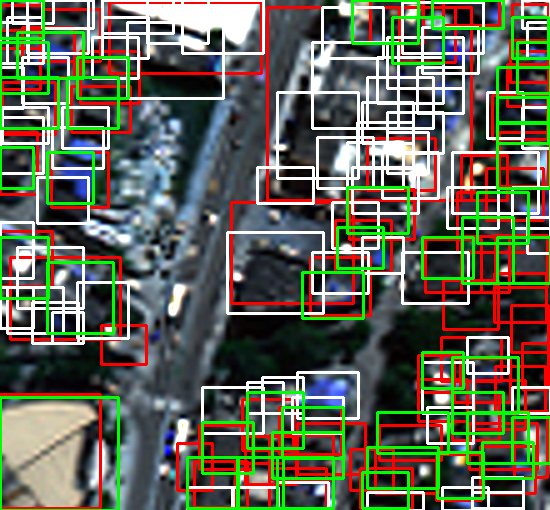}}}\\
		13-bit LR 3-band & \Vcentre{\subfloat{\includegraphics[width = 1.6in]{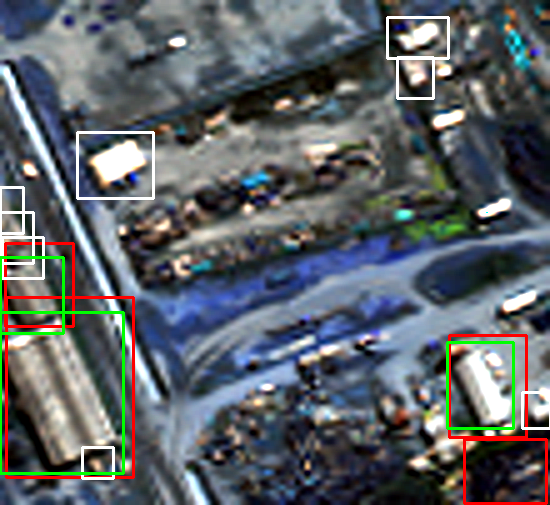}}} &
		\Vcentre{\subfloat{\includegraphics[width = 1.6in]{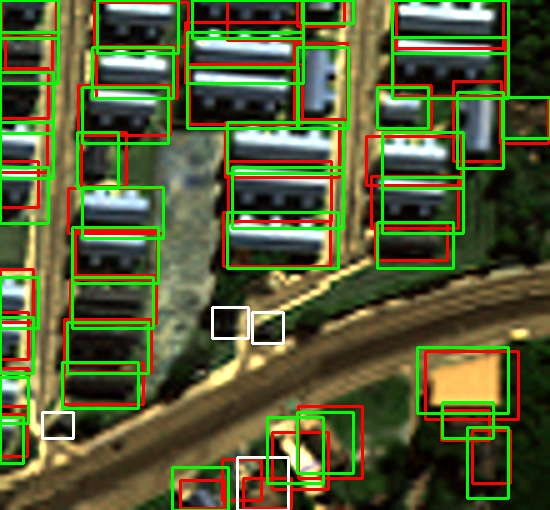}}} &
		\Vcentre{\subfloat{\includegraphics[width = 1.6in]{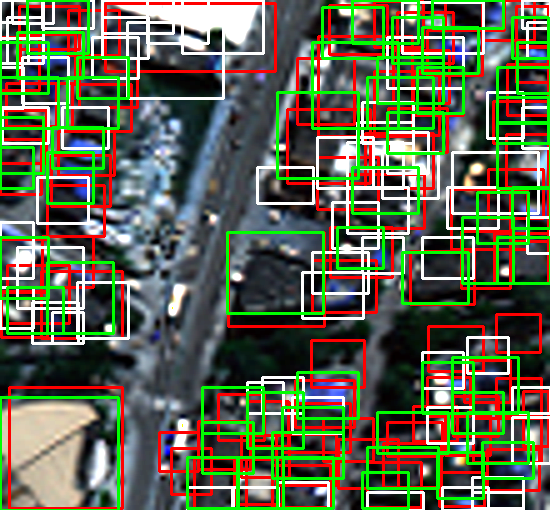}}}\\
		13-bit LR 3-band (false color) & \Vcentre{\subfloat{\includegraphics[width = 1.6in]{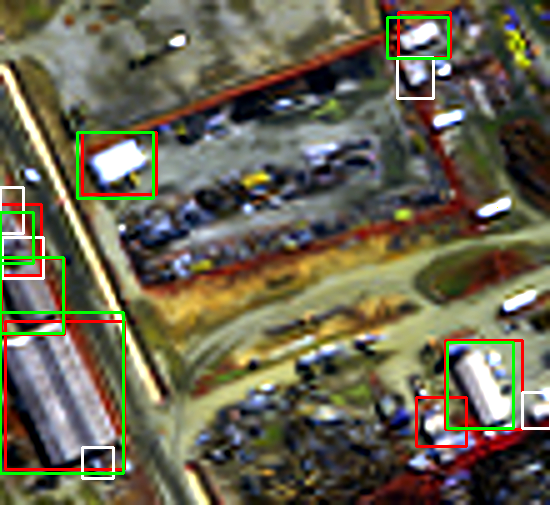}}} &
		\Vcentre{\subfloat{\includegraphics[width = 1.6in]{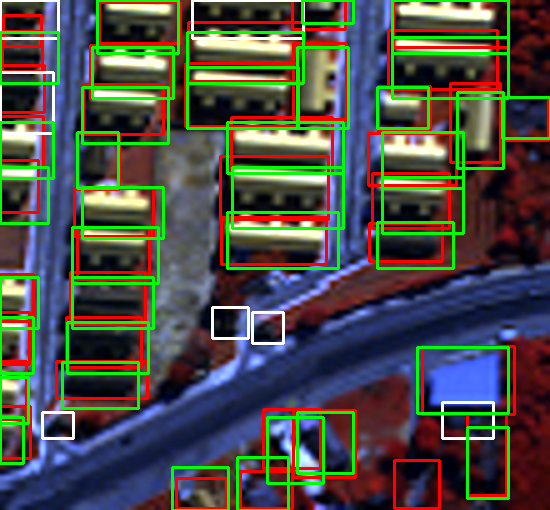}}} &
		\Vcentre{\subfloat{\includegraphics[width = 1.6in]{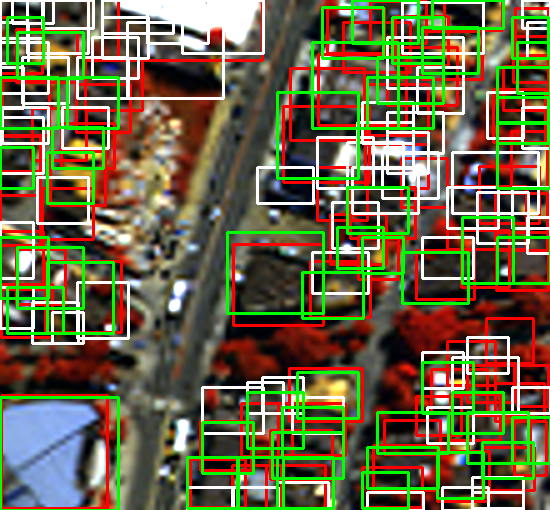}}}\\
		13-bit LR 8-band & \Vcentre{\subfloat{\includegraphics[width = 1.6in]{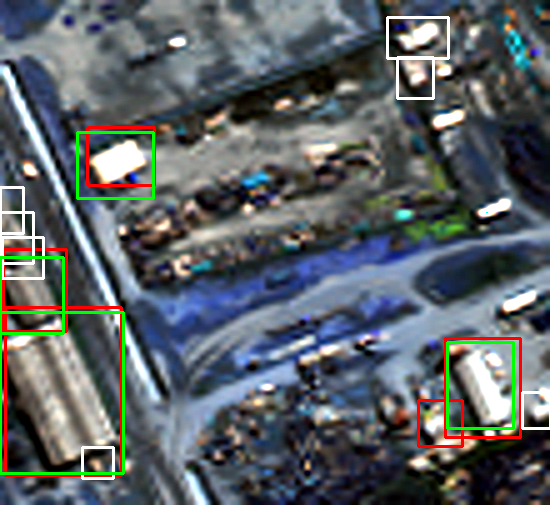}}} &
		\Vcentre{\subfloat{\includegraphics[width = 1.6in]{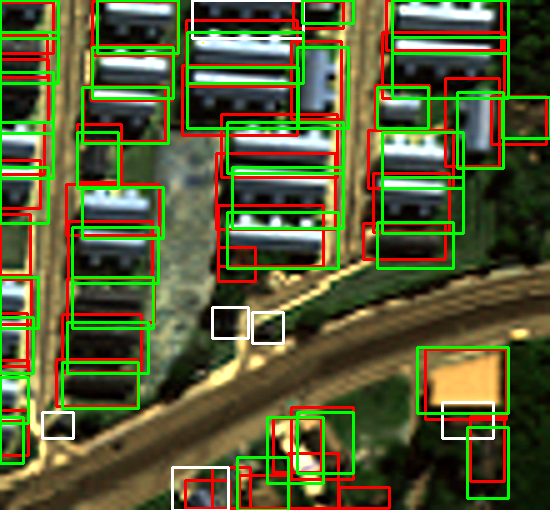}}} &
		\Vcentre{\subfloat{\includegraphics[width = 1.6in]{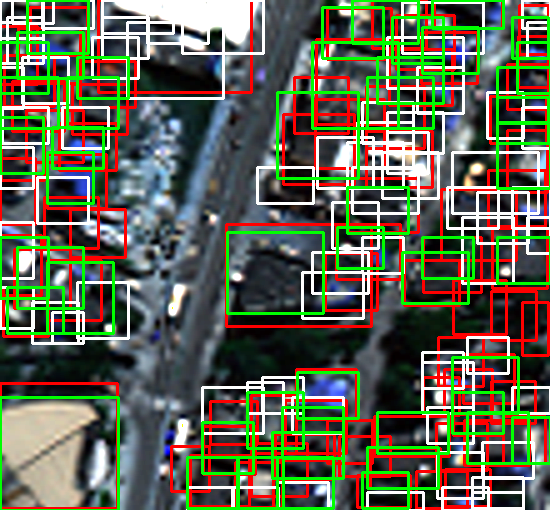}}}
	\end{tabular}
	\caption{Predicted R-FCN with OHEM examples for high density, moderate density, and low density scenes. Predicted bounding boxes are shown in red, ground truth bounding boxes not matched to a true prediction are in white, while green bounding boxes represent true positives at an IoU threshold of 0.5. For the purpose of display, only the standard RGB bands are shown in the 8-band row. All imagery in this figure is from DigitalGlobe.}
	\label{fig:density_visual}
\end{figure*}

Due to the large variation in scene density and intraclass object size inherent in SpaceNet, it is informative to evaluate performance as a function of these parameters. Subplot (A) in Figure \ref{fig:density} highlights detector performance broken down by density level within the test scenes. As enumerated in Figure \ref{fig:densitydist} (A), scenes with low density include fewer than 40 buildings, while high density scenes are categorized as scenes with more than 90 buildings which is on par with residential and suburban neighborhoods. Scenes that include between 40 and 90 buildings fall in the moderate density category. As expected, performance drops for low density scenes. We attribute this to the fact that when a scene includes a small number of buildings, if even one or two of the targets are missed by the detector, then recall takes a large hit, resulting in a lower $F_1$ Score. The slight drop in performance from moderate to high density scenes is likely due to documented intrinsic weaknesses in state-of-the-art-detectors in highly cluttered scenes where objects tend to be small \cite{DBLP:journals/corr/LiuAESR15,aerial,DSSDpaper}. Some qualitative examples are presented in Figure \ref{fig:density_visual}.

Figure \ref{fig:size}, subplot (B) describes detector performance for very small, small, medium, large, and very large building categories (as defined in Figure \ref{fig:sizedist} (B)) at 0.5 IoU and 0.5 detector confidence threshold. These buildings can cover the entire footprint of an annotated scene (over $40,000m^2$) or can be as small as $4\times4$ pixels in the LR case. As was the case in the overall performance and density study, here, too, we note similar behaviors amongst the relationship of resolution, dynamic range and spectral information. Also, as expected, we note a steady performance improvement as the buildings become larger. Interestingly, $F_1$ Scores for the very small and very large building categories extracted from the high bit-depth, LR, resized images are statistically significant when compared to the lower bit-depth, HR counterparts. Table \ref{tab:IoU_SizeStudy} provides more insight into the relationship between performance and IoU thresholds. In particular, when considering small buildings, one may want to relax the IoU threshold, considering often times there are only a few pixels on target in these instances. This boost in performance when relaxing the IoU threshold to 0.2 from 0.5 is most significant for very small buildings and seems to taper off in the case of the other categories.
\begin{table}
	\centering
	\small
	\begin{tabular}{ccrrrr}
		\toprule
		\multicolumn{3}{c}{\multirow{2}[2]{*}{$F_1$ Score}} & \multicolumn{3}{c}{Confidence Threshold} \\
		\multicolumn{3}{c}{} & 0.2   & 0.5   & 0.75 \\
		\cmidrule{4-6}    \multirow{20}[10]{*}{\begin{sideways}IoU Threshold\end{sideways}} & \multirow{4}[2]{*}{\begin{sideways}Very Small\end{sideways}} & 0.2   & \cellcolor[rgb]{ .996,  .871,  .506} 0.536 & \cellcolor[rgb]{ .996,  .867,  .506} 0.533 & \cellcolor[rgb]{ .973,  .435,  .424} 0.125 \\
		&       & 0.3   & \cellcolor[rgb]{ .992,  .827,  .498} 0.496 & \cellcolor[rgb]{ .992,  .827,  .498} 0.497 & \cellcolor[rgb]{ .973,  .431,  .424} 0.123 \\
		&       & 0.4   & \cellcolor[rgb]{ .988,  .769,  .486} 0.439 & \cellcolor[rgb]{ .988,  .773,  .486} 0.444 & \cellcolor[rgb]{ .973,  .427,  .42} 0.117 \\
		&       & 0.5   & \cellcolor[rgb]{ .984,  .663,  .467} 0.340 & \cellcolor[rgb]{ .984,  .667,  .467} 0.346 & \cellcolor[rgb]{ .973,  .412,  .42} 0.102 \\
		\cmidrule{2-6}          & \multirow{4}[2]{*}{\begin{sideways}Small\end{sideways}} & 0.2   & \cellcolor[rgb]{ .918,  .898,  .514} 0.623 & \cellcolor[rgb]{ .863,  .882,  .51} 0.651 & \cellcolor[rgb]{ .98,  .596,  .455} 0.278 \\
		&       & 0.3   & \cellcolor[rgb]{ .98,  .918,  .518} 0.593 & \cellcolor[rgb]{ .918,  .898,  .514} 0.624 & \cellcolor[rgb]{ .98,  .592,  .455} 0.275 \\
		&       & 0.4   & \cellcolor[rgb]{ .996,  .89,  .51} 0.555 & \cellcolor[rgb]{ .996,  .922,  .518} 0.587 & \cellcolor[rgb]{ .98,  .588,  .451} 0.269 \\
		&       & 0.5   & \cellcolor[rgb]{ .992,  .808,  .494} 0.478 & \cellcolor[rgb]{ .992,  .839,  .502} 0.507 & \cellcolor[rgb]{ .98,  .569,  .447} 0.252 \\
		\cmidrule{2-6}          & \multirow{4}[2]{*}{\begin{sideways}Medium\end{sideways}} & 0.2   & \cellcolor[rgb]{ .765,  .855,  .506} 0.699 & \cellcolor[rgb]{ .722,  .843,  .502} 0.718 & \cellcolor[rgb]{ .988,  .722,  .478} 0.395 \\
		&       & 0.3   & \cellcolor[rgb]{ .824,  .871,  .51} 0.670 & \cellcolor[rgb]{ .776,  .859,  .506} 0.691 & \cellcolor[rgb]{ .988,  .718,  .478} 0.392 \\
		&       & 0.4   & \cellcolor[rgb]{ .898,  .894,  .514} 0.633 & \cellcolor[rgb]{ .851,  .878,  .51} 0.656 & \cellcolor[rgb]{ .988,  .714,  .475} 0.387 \\
		&       & 0.5   & \cellcolor[rgb]{ .996,  .89,  .51} 0.556 & \cellcolor[rgb]{ .996,  .914,  .514} 0.577 & \cellcolor[rgb]{ .984,  .69,  .471} 0.368 \\
		\cmidrule{2-6}          & \multirow{4}[2]{*}{\begin{sideways}Large\end{sideways}} & 0.2   & \cellcolor[rgb]{ .565,  .796,  .494} 0.795 & \cellcolor[rgb]{ .537,  .788,  .494} 0.809 & \cellcolor[rgb]{ .996,  .847,  .502} 0.515 \\
		&       & 0.3   & \cellcolor[rgb]{ .612,  .812,  .498} 0.771 & \cellcolor[rgb]{ .58,  .8,  .494} 0.787 & \cellcolor[rgb]{ .992,  .847,  .502} 0.513 \\
		&       & 0.4   & \cellcolor[rgb]{ .678,  .831,  .502} 0.739 & \cellcolor[rgb]{ .643,  .82,  .498} 0.757 & \cellcolor[rgb]{ .992,  .839,  .502} 0.508 \\
		&       & 0.5   & \cellcolor[rgb]{ .82,  .871,  .51} 0.672 & \cellcolor[rgb]{ .784,  .859,  .506} 0.689 & \cellcolor[rgb]{ .992,  .82,  .498} 0.487 \\
		\cmidrule{2-6}          & \multirow{4}[2]{*}{\begin{sideways}Very Large\end{sideways}} & 0.2   & \cellcolor[rgb]{ .427,  .757,  .486} 0.861 & \cellcolor[rgb]{ .388,  .745,  .482} 0.879 & \cellcolor[rgb]{ .863,  .882,  .51} 0.651 \\
		&       & 0.3   & \cellcolor[rgb]{ .471,  .773,  .49} 0.839 & \cellcolor[rgb]{ .427,  .757,  .486} 0.860 & \cellcolor[rgb]{ .875,  .886,  .514} 0.646 \\
		&       & 0.4   & \cellcolor[rgb]{ .569,  .8,  .494} 0.793 & \cellcolor[rgb]{ .525,  .784,  .49} 0.814 & \cellcolor[rgb]{ .914,  .898,  .514} 0.625 \\
		&       & 0.5   & \cellcolor[rgb]{ .769,  .855,  .506} 0.697 & \cellcolor[rgb]{ .725,  .843,  .502} 0.718 & \cellcolor[rgb]{ .996,  .918,  .514} 0.580 \\
		\bottomrule
	\end{tabular}%
	\caption[Detector IoU comparison]{Comparision of per building size category $F_1$ Score resulting from different combinations of confidence and IoU thresholds for the LR, 13-bit, RGB, rescaled data.}
	\label{tab:IoU_SizeStudy}%
\end{table}%
\section{Discussion}
\label{sec:discussion}
This paper presents a baseline study for one of the most efficient and accurate object detectors when applied to satellite images. We provided insight on how additional information, such as higher dynamic range or more spectral bands, affects performance for variable object size and scene density.  We showed that comparable performance between data of different GSDs is attainable if additional bit-depth is exploited in lower resolution images. We also note that R-FCN performs poorly in scenes with a high object density, which corroborated observations by the authors of the SSD and YOLO detectors on ground-based images. Considering the use of out-of-the-box parameters, it is not surprising that very small buildings present a challenge for these detectors.  Tuning the hyperparameters, adjusting the RPN anchors and scales, or using earlier activation layers (as suggested by \cite{COWCpaper}) could yield better optimized results for this category.

Despite these challenges, our analysis demonstrates the importance of leveraging all available data. Such observations could provide researchers an edge in the SpaceNet challenge, where the goal is to find automated methods for extracting map-ready building footprints. As we have shown, this challenging dataset, with its widely varying building footprints, may require different detectors for different scenarios. For example, specialized residential or commercial building detectors may outperform a single all-encompassing model. Our task in this study focused on bounding boxes instead of footprints with the goal of harnessing the additional information from the dynamic range and spectral bands to aid in detection tasks and improve performance that is on par with the winners of the initial SpaceNet challenge.

Future studies could incorporate more scene geographic diversity, object classes, and detector architectures. These studies will assess the generalizability of detectors on satellite imagery, an important next step for this community. 

\bibliographystyle{ieee}
\bibliography{detector_overhead_bib2}
\end{document}